\documentclass[conference]{IEEEtran}
\pdfoutput=1
\IEEEoverridecommandlockouts
\usepackage{comment}
\usepackage{cite}
\usepackage{amsmath,amssymb,amsfonts}
\usepackage{algorithmic}
\usepackage{graphicx}
\usepackage{textcomp}
\usepackage{xcolor}
\usepackage{booktabs}
\usepackage{wrapfig}
\def\BibTeX{{\rm B\kern-.05em{\sc i\kern-.025em b}\kern-.08em
    T\kern-.1667em\lower.7ex\hbox{E}\kern-.125emX}}
\begin{document}

\title{APPFLx: Providing Privacy-Preserving Cross-Silo Federated Learning as a Service
}
\author{\IEEEauthorblockN{Zilinghan Li\IEEEauthorrefmark{1}\IEEEauthorrefmark{2}\IEEEauthorrefmark{3},
Shilan He\IEEEauthorrefmark{1}\IEEEauthorrefmark{3}\IEEEauthorrefmark{4},
Pranshu Chaturvedi\IEEEauthorrefmark{1}\IEEEauthorrefmark{2}\IEEEauthorrefmark{3}, 
Trung-Hieu Hoang\IEEEauthorrefmark{4},
Minseok Ryu\IEEEauthorrefmark{5},
E. A. Huerta\IEEEauthorrefmark{1}\IEEEauthorrefmark{6}\IEEEauthorrefmark{7},\\
Volodymyr Kindratenko\IEEEauthorrefmark{2}\IEEEauthorrefmark{3}\IEEEauthorrefmark{4},
Jordan Fuhrman \IEEEauthorrefmark{1}\IEEEauthorrefmark{9},
Maryellen Giger \IEEEauthorrefmark{1}\IEEEauthorrefmark{9},
Ryan Chard\IEEEauthorrefmark{1}\IEEEauthorrefmark{8},
Kibaek Kim\IEEEauthorrefmark{9},
Ravi Madduri\IEEEauthorrefmark{1}\IEEEauthorrefmark{8}
}

\IEEEauthorblockA{\IEEEauthorrefmark{1}Data Science and Learning Division, Argonne National Laboratory, Lemont, IL 60439 USA}
\IEEEauthorblockA{\IEEEauthorrefmark{2}Department of Computer Science, University of Illinois at Urbana-Champaign, Urbana, IL 61801 USA}
\IEEEauthorblockA{\IEEEauthorrefmark{3}National Center for Supercomputing Applications, University of Illinois at Urbana-Champaign, Urbana, IL 61801 USA}
\IEEEauthorblockA{\IEEEauthorrefmark{4}Department of Electrical and Computer Engineering, University of Illinois at Urbana-Champaign, Urbana, IL 61801 USA}
\IEEEauthorblockA{\IEEEauthorrefmark{5}School of Computing and Augmented Intelligence, Arizona State University, Tempe, AZ 85281 USA}
\IEEEauthorblockA{\IEEEauthorrefmark{6}Department of Physics, University of Illinois at Urbana-Champaign, Urbana, IL 61801 USA}
\IEEEauthorblockA{\IEEEauthorrefmark{7}Department of Computer Science, The University of Chicago, Chicago, IL 60637 USA}
\IEEEauthorblockA{\IEEEauthorrefmark{8}University of Chicago Consortium for Advanced Science and Engineering, Chicago, IL 60637 USA}
\IEEEauthorblockA{\IEEEauthorrefmark{9}Mathematics and Computer Science Division, Argonne National Laboratory, Lemont, IL 60439 USA}
\IEEEauthorblockA{\IEEEauthorrefmark{10}Department of Radiology, University of Chicago, Hyde Park, IL 60616 USA}
}


\maketitle

\begin{abstract}
Cross-silo privacy-preserving federated learning (PPFL) is a powerful tool to collaboratively train robust and generalized machine learning (ML) models 
without sharing sensitive (e.g., healthcare of financial) local data. To ease and accelerate the adoption of 
PPFL, we introduce \textit{APPFLx}, a ready-to-use 
platform that provides privacy-preserving cross-silo federated learning as a service. \textit{APPFLx} employs Globus authentication to allow users to easily and securely invite trustworthy collaborators for PPFL, implements several synchronous and asynchronous FL algorithms, streamlines the FL experiment launch process, and enables tracking and visualizing the life cycle of FL experiments, allowing domain experts and ML practitioners to easily orchestrate and evaluate cross-silo FL under one platform. \textit{APPFLx} is available online at https://appflx.link
\end{abstract}


\section{Introduction}\label{sec:intro}
Building robust machine learning (ML) models that are resilient to domain shift~\cite{finlayson2021clinican} requires training across diverse, oftentimes private, datasets; as well as access to large computing 
resources. Federated learning (FL)~\cite{fedavg} is a collaborative learning approach, capable of addressing domain shift challenges, where multiple data owners, referred to as clients, train a model together under the orchestration of a central server by sharing the ML model trained on their local datasets instead of sharing the data directly. 

FL enables the creation of more robust models without the exposure of local datasets. However, FL by itself, does not guarantee the privacy of data, because the information extracted from the communication of FL algorithms can be accumulated and utilized to infer the private local data used for training~\cite{Hatamizadeh_2022_CVPR}. To address these challenges, we developed an open-source software framework, called Argonne Privacy Preserving Federated Learning (APPFL)~\cite{appfl}, whose algorithmic advances in differential privacy 
enable privacy-preserving federated learning (PPFL). 
APPFL enables the training of ML models in a distributed setting across multiple institutions, where sensitive data are located, with the ability to scale on supercomputing resources. With APPFL, 
researchers can develop robust, trustworthy ML models 
for applications in biomedicine and smart grid applications, where data privacy is essential. 


Setting up a secure FL experiment that needs high-performance computing resources across distributed sites requires technical capabilities that may not be available for all. To lower the technical and cybersecurity barriers to entry for leveraging PPFL, and to enable domain experts in large institutions to utilize FL, we created the Argonne Privacy-Preserving Federated Learning as a service (\textit{APPFLx}), which enables cross-silo PPFL with user-friendly web interfaces for managing, deploying, analyzing, and visualizing PPFL experiments. 
\textit{APPFLx} also enables the creation of secure federations with end-to-end strong Identity and Access Management (IAM), where collaborators across organizational boundaries can create a new federation or join an existing federation using their institutional identities, perform training on datasets at their respective institutions and securely share the privacy-preserving model weights with the service to enable secure aggregation. 

Existing PPFL frameworks typically involve downloading and configuring complex software, manually creating trust boundaries and identities to enable gradient aggregation, and understanding of technical details of the underlying deep learning software stack to enable distributed training which is cumbersome~\cite{beutel2020flower}\cite{roth2022nvidia}. 
In stark contrast, \textit{APPFLx} features include: 
1) secure distributed training on heterogeneous computing resources, along with over half a dozen federation strategies; 2) both synchronous and asynchronous aggregation; 3) integration with TensorBoard capabilities; 4) interfaces to examine data distributions and resource utilization across different sites; 5) detailed reports of different experiments, ability to use model architectures from GitHub or pre-trained models from HuggingFace model repository; and 6) the ability to set different hyper-parameters of the experiments (like privacy budget to be used in training) from the convenient web interface. Additionally, we developed comprehensive approaches to measure privacy protection by attacking models generated by \textit{APPFLx} and implemented FAIR standards~\cite{2022NatSD...9..657R,Huerta_2023_FAIRAI} for capturing and storing the metadata for all the PPFL experiments to 
ensure reproducibility. 

\section{System Overview}\label{sec:overview}

\begin{figure}[!htbp]
\centering
\includegraphics[width=0.48\textwidth]{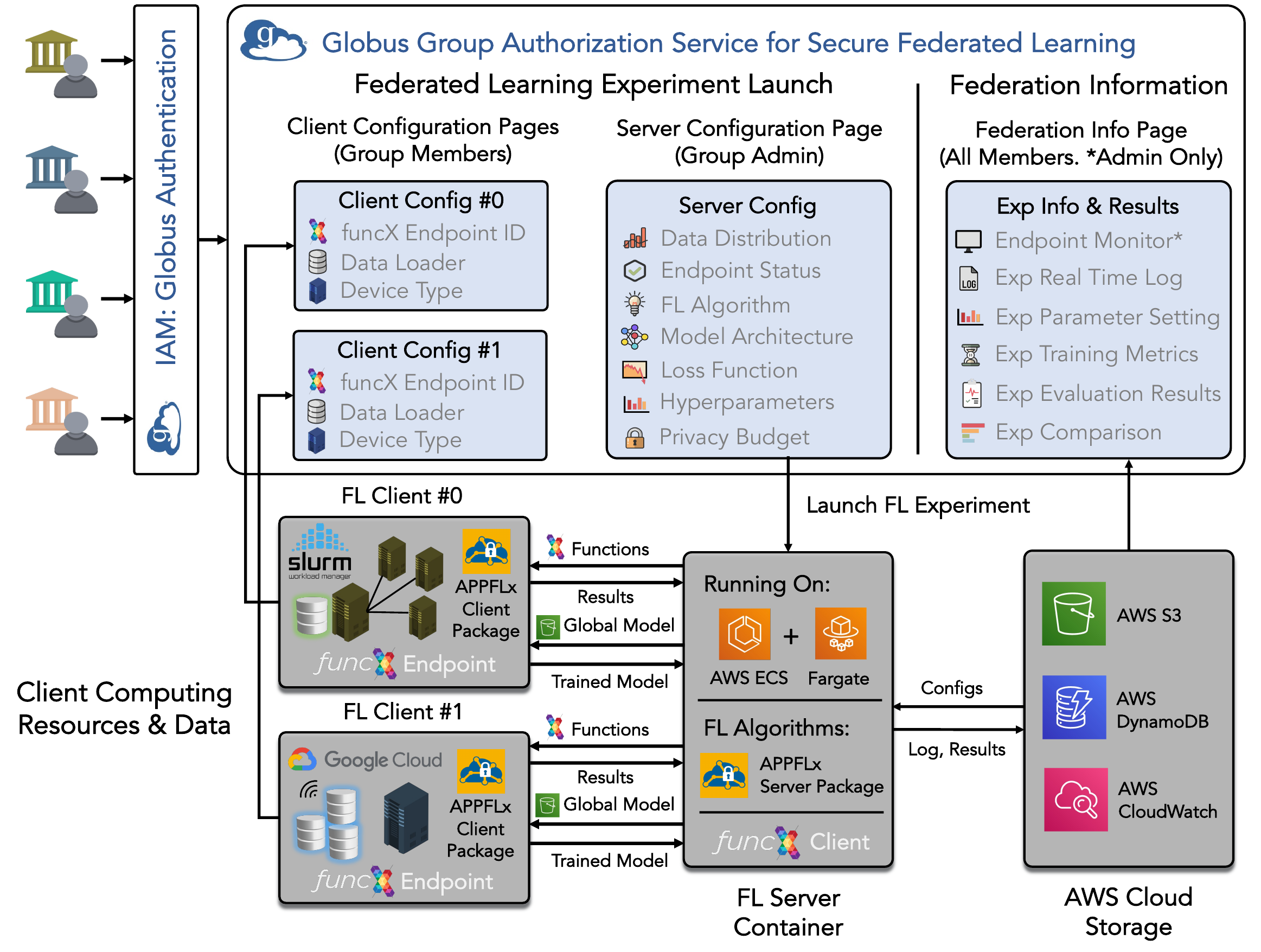} 
\caption{System overview of \textit{APPFLx}.}
\label{fig:overview}
\end{figure}

In this section we describe the main components of \textit{APPFLx}; see Figure~\ref{fig:overview} for the system overview.

\textbf{Identity and Access Management (IAM).} In cross-silo PPFL, ensuring the verification of collaborating clients' identities is of utmost importance to prevent potential attacks from Byzantine clients \cite{lamport2019byzantine}. \textit{APPFLx} utilizes the Globus authentication service \cite{globus, saas, globusauth} for user identification. Authenticated users can create a ``federation'' using the Globus group authorization service and extend invitations to trusted collaborators, allowing them to conduct FL experiments securely and efficiently.

\textbf{Client Computing Resources \& Data.} In an FL group, each client has access to private local datasets. The composition of FL clients is generally heterogeneous in computing capabilities, and each FL client may also have a heterogeneous computing resource. Examples of typical FL clients include high performance computing clusters with job schedulers such as SLURM and cloud computing machines. \textit{APPFLx} allows users to easily register their heterogeneous computing resources for FL by following a simple one-time setup. This setup process includes creating a Globus Compute (previously known as funcX) \cite{funcx} endpoint and installing the APPFLx client package on the computing resource. FuncX is a distributed Function as a Service (FaaS) platform used by \textit{APPFLx} to dispatch the training task codebase from the server to clients for local training. Clients are required to open ports 443 (HTTPS) and 5671 (AMQPS) for outbound traffic to transmit tasks and results between the funcX service and the endpoints, where HTTPS is used for retrieving credentials and AMQPS is used for encrypted task/result transmission. The APPFLx client package contains the codebase that provides seamless support for remote training tasks. The process of setting up a funcX endpoint also incorporates Globus authentication, ensuring that only members belonging to the same FL group are authorized to send codebase for execution on the funcX endpoints. This stringent security measure fosters a protected environment for the FL collaborators.

\textbf{Client Configuration Page.} Once the clients have created funcX endpoints, they can register their computing resources by providing the funcX endpoint ID and the device type on the client configuration page of \textit{APPFLx}. Furthermore, clients are required to upload a data loader file on this page that enables the dispatched training codebase to access their local data on their computing resources during the training. 

\textbf{Server Configuration Page.} The FL group administrator is responsible for configuring the FL settings on the \textit{APPFLx} server configuration page. These settings include the FL algorithm, training model architecture, training loss function, hyper-parameters (e.g., learning rate, number of local and global epochs), and privacy budgets. In particular, \textit{APPFLx} provides five synchronous FL algorithms (\texttt{FedAvg} \cite{fedavg}, \texttt{FedAvgM} \cite{fedavgm}, \texttt{FedAdam}, \texttt{FedAdagrad}, and \texttt{FedYogi} \cite{fedadaptive}) and two asynchronous FL algorithms (\texttt{FedAsync} \cite{fedasync} and \texttt{FedBuf} \cite{fedbuf}). For the training model, \textit{APPFLx} offers flexible choices --- users can choose pre-defined template models or upload their own model definition files from their local computers. Moreover, the platform seamlessly integrates with GitHub, promoting enhanced collaboration opportunities among the FL group members, and it also integrates with HuggingFace to allow users to leverage pre-trained models, further enhancing the efficiency of the FL. Understanding training data distribution across the different sites in a federation is an important aspect to reason with FL model performance. On the server configuration page, the FL administrator can visualize the distribution of the training data across available sites before starting the FL process.  Additionally, the FL administrator can check the status of client endpoints and launch an FL experiment on this page.

\textbf{FL Server Container.} Upon initiation by the FL group administrator, \textit{APPFLx} launches an FL orchestration server in a container using AWS Elastic Container Service (ECS) and AWS Fargate, facilitating both scalability and robustness. The server retrieves the experiment configurations from AWS S3 and acts as a funcX client to dispatch training functions to the client funcX endpoints to train models using their local data. \textit{APPFLx} employs the AWS S3 bucket for model transfers between the server container and client endpoints. The server aggregates the locally trained models based on the choice of the FL algorithms available in APPFL.
As the FL experiment unfolds, the server stores all the experiment logs and results in the AWS cloud storage, ensuring comprehensive and reliable record-keeping.

\textbf{AWS Cloud Storage.} \textit{APPFLx} utilizes AWS cloud storage services to securely and reliably store all FL experiment information. Specifically, real-time training logs are seamlessly managed by AWS CloudWatch. DynamoDB is used to store experiment metadata, and AWS S3 plays a pivotal role in storing a diverse array of experiment-related files, including configuration files, various result files, and training log backups. Incorporating these AWS cloud storage services, \textit{APPFLx} guarantees a resilient infrastructure, allowing users to confidently engage in FL experiments while maintaining the utmost data integrity and privacy.

\textbf{Federation Information Page.} Powered by the AWS cloud storage, \textit{APPFLx} provides an information page accessible to all FL group members. This page provides a wealth of valuable information regarding the FL experiments conducted within the group. Users can access real-time training logs for immediate monitoring and insightful analysis of ongoing training processes. Experiment configurations are also available for review on this page. In addition, users can access essential training metrics, such as the time taken for each client to complete specific training tasks, for in-depth performance evaluation. Furthermore, the page provides reports containing the local validation accuracy results for each client. \textit{APPFLx} can also generate a comparison report, which empowers users to conduct thorough comparisons across multiple experiment runs, facilitating data-driven decision-making and fostering continuous improvements in the FL process. Finally, the group administrator is provided with an endpoint information section to monitor the CPU, GPU, memory, and network usage of the client endpoints in real-time. The federation information page serves as an invaluable resource, promoting collaboration, transparency, and efficiency within the FL group.


\section{Experiment}
\subsection{Experiment Settings and Launch}
In this section, we present a use case that employs \textit{APPFLx} to conduct federated learning with a group of five clients to train a convolutional neural network (CNN) with two convolutional layers using the artificially partitioned MNIST datasets \cite{mnist}. The MNIST dataset is partitioned equally into five chunks as the local datasets of the five FL clients. Two synchronous FL algorithms, \texttt{FedAvg} \cite{fedavg} and \texttt{FedAvgM} \cite{fedavgm}, are employed to train the model. The FL training takes 10 global communication rounds, and each client performs 2 local epochs in each round with batch size 64 and local learning rate 0.01. Local learning rate is decayed by a factor of 0.975 for each round, and the server momentum for $\texttt{FedAvgM}$ is equal to 0.9.
Table \ref{tab:endpoint} shows the information of five heterogeneous client endpoints. Notably, all the endpoints only use the CPU for training. The group members use the client configuration page to register their endpoints and upload data loaders for their local datasets. The group administrator launches the FL experiment by specifying the training hyper-parameters in the server configuration page.

\begin{table}[!htbp]
  \centering
  \setlength{\tabcolsep}{6pt}
  \setlength{\belowcaptionskip}{2mm}
  \caption{Client endpoints information.} \label{tab:endpoint}	
    \begin{tabular}{ll}
		\toprule  
		Endpoint name & Machine\\
		\midrule 
        \texttt{delta-cpu-01} & NCSA Delta supercomputer\\
        \texttt{delta-gpu-01} & NCSA Delta supercomputer\\
        \texttt{mydefconf} & ALCF Polaris supercomputer\\
        \texttt{crn-azure} & Microsoft Azure virtual machine\\
        \texttt{appfl-test} & MacBook Pro, 2021, M1 Chip\\
		\bottomrule 
	\end{tabular}
\end{table}
\vspace{-5pt}


\subsection{Experiment Information and Results}
\begin{figure}[!htbp]
\centering
\includegraphics[width=0.98\columnwidth]{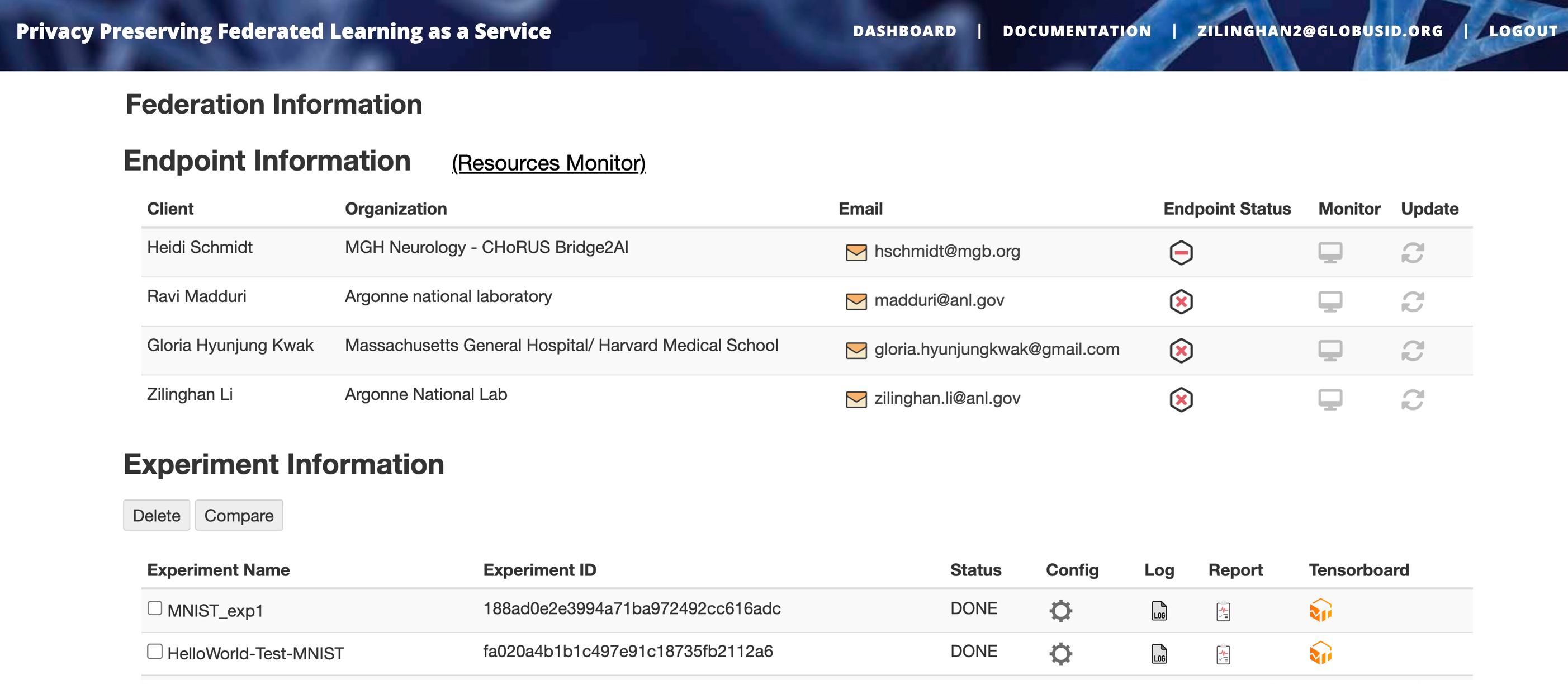}
\caption{Federation information page.}
\label{fig:federation_info}
\end{figure}

\begin{figure}[!htbp]
\centering
\includegraphics[width=0.98\columnwidth]{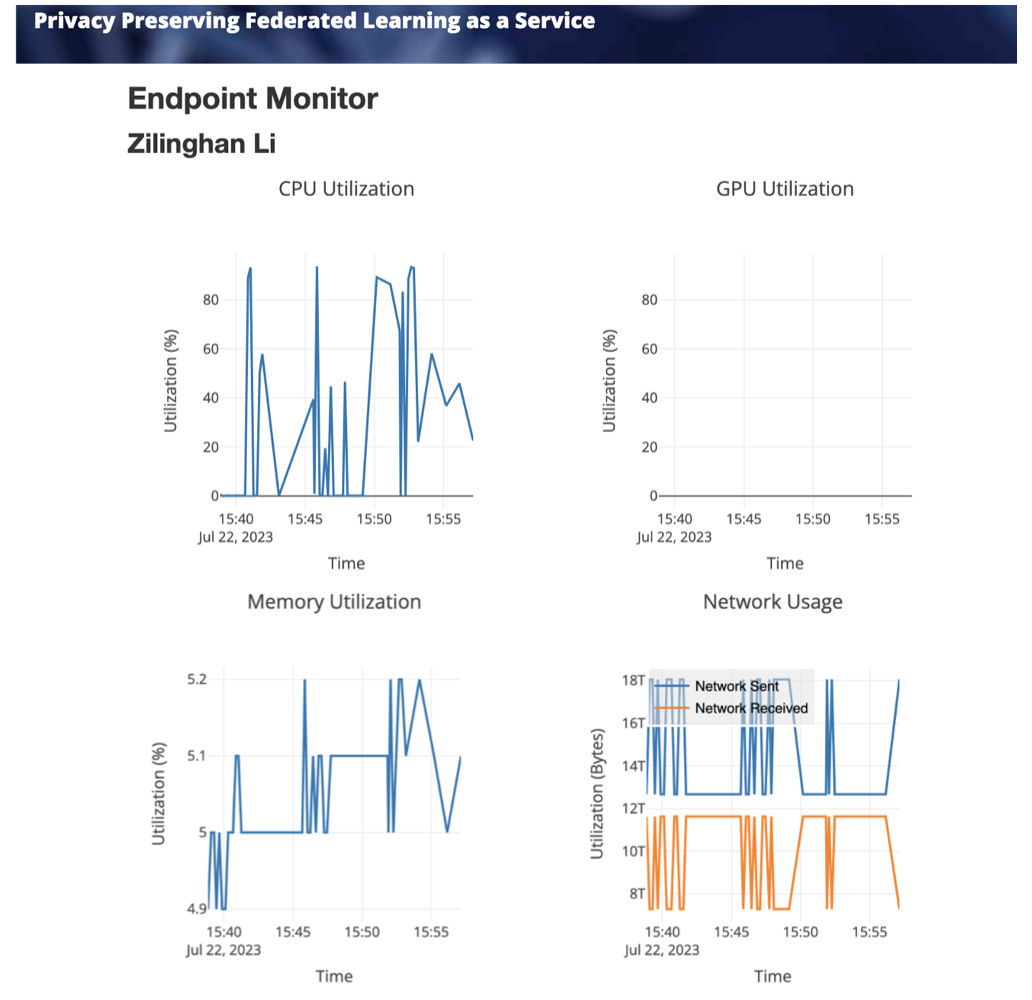}
\caption{Client endpoints monitor page.}
\label{fig:endpoint_monitor}
\end{figure}

\begin{figure}[!htbp]
\centering
\includegraphics[width=0.98\columnwidth]{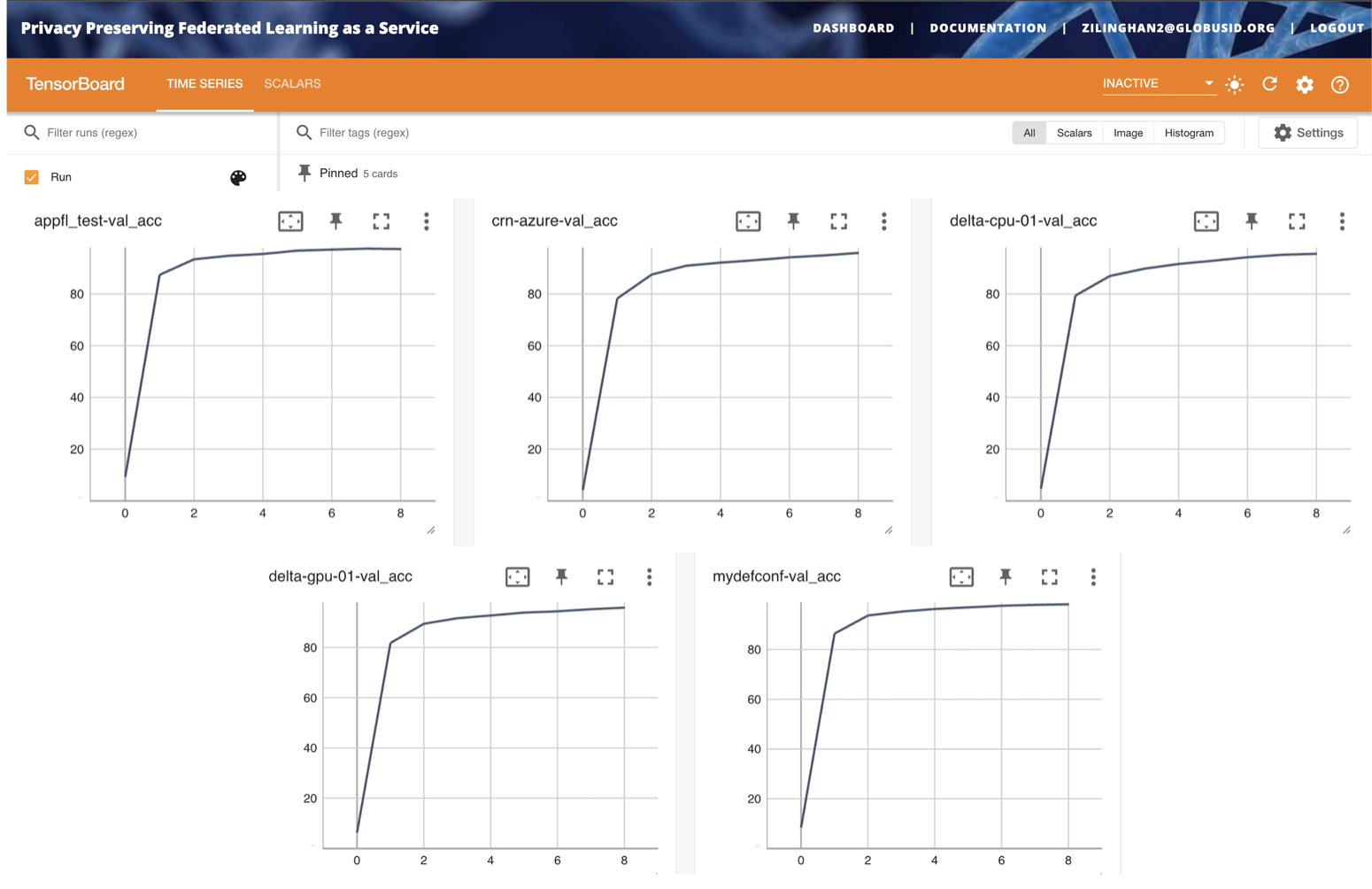}
\caption{Tensorboard visualization page showing ML model performance at five different sites.}
\label{fig:tensorboard}
\end{figure}

\begin{figure}[!htbp]
\centering
\includegraphics[width=0.98\columnwidth]{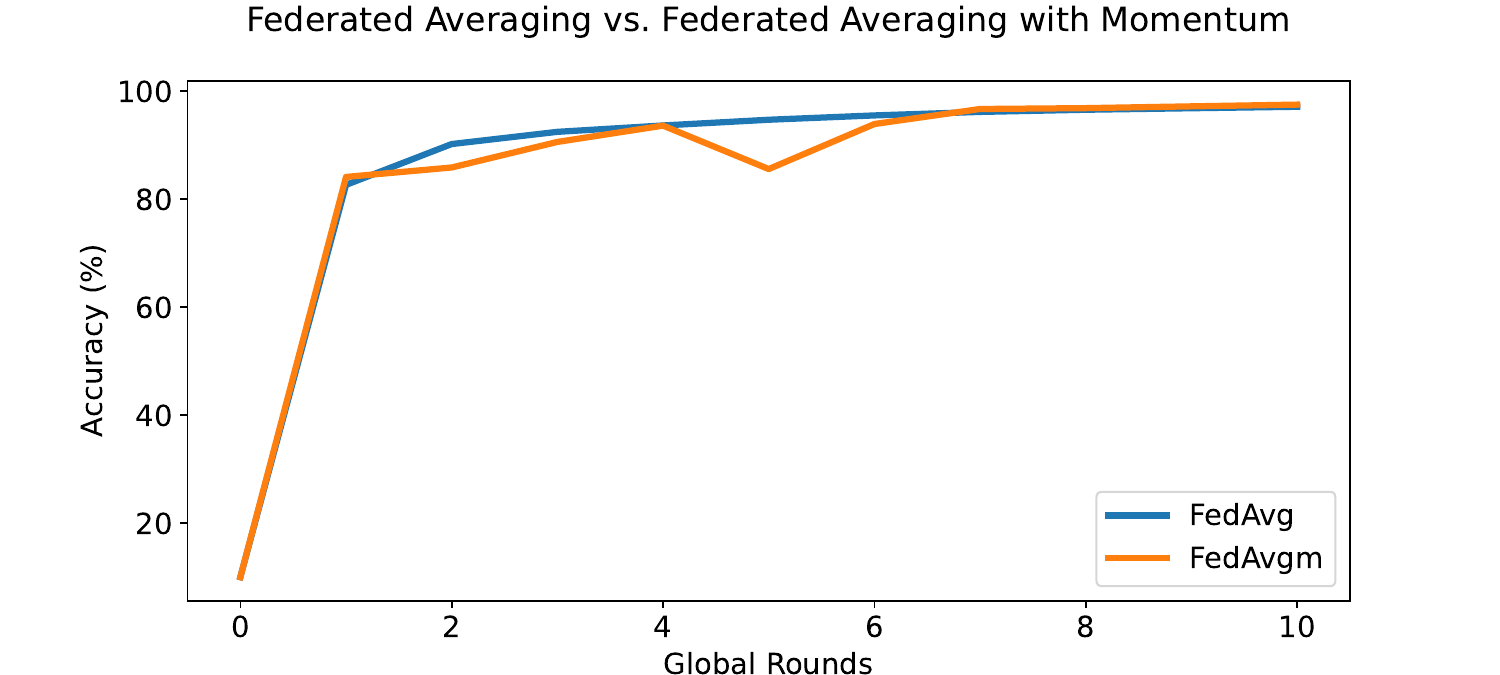}
\caption{Validation accuracy of \texttt{FedAvg} and \texttt{FedAvgM} during the federated learning process.}
\label{fig:accuracy}
\end{figure}

As previously mentioned, \textit{APPFLx} offers a federation information page for all group members to access the information and results of the FL experiments, and the group administrator is also provided with an endpoint information section to monitor the status of all the client endpoints, as shown in Figure \ref{fig:federation_info}. Figure \ref{fig:endpoint_monitor} showcases the client endpoint monitor page, presenting real-time resource utilization for each client endpoint. For launched FL experiments, group members have access to various components, including real-time logs, experiment configurations, experiment reports, and tensorboard visualizations (Figure \ref{fig:tensorboard}). Figure \ref{fig:accuracy} shows the change in the validation accuracy of \texttt{FedAvg} and \texttt{FedAvgM} algorithms during the training process.

\section{Conclusion}
\textit{APPFLx} is a user-friendly platform that provides privacy-preserving cross-silo federated learning as a service. With streamlined FL experiment launching process and comprehensive experiment results, \textit{APPFLx} enhances the accessibility and ease of use of privacy-preserving federated learning for professionals across diverse industries. This efficient and collaborative environment empowers users to develop robust and generalized ML models securely, promoting advancements in various fields while safeguarding data privacy.

\section*{Acknowledgments}
The material is based upon work and resources of the Argonne Leadership Computing Facility at Argonne National Laboratory, which is supported by the U.S. Department of Energy, Office of Science, under contract number DE-AC02-06CH11357 and in part by MIDRC (The Medical Imaging and Data Resource Center) funded by the the National Institutes of Health under contract 75N92020C00021.
This research is also part of the Delta research computing project, which is supported by the National Science Foundation (award OCI 2005572), and the State of Illinois. Delta is a joint effort of the University of Illinois at Urbana-Champaign and its National Center for SuperComputing Applications.

\bibliographystyle{IEEEtran}
\bibliography{output}

\begin{thebibliography}{10}
\providecommand{\url}[1]{#1}
\csname url@samestyle\endcsname
\providecommand{\newblock}{\relax}
\providecommand{\bibinfo}[2]{#2}
\providecommand{\BIBentrySTDinterwordspacing}{\spaceskip=0pt\relax}
\providecommand{\BIBentryALTinterwordstretchfactor}{4}
\providecommand{\BIBentryALTinterwordspacing}{\spaceskip=\fontdimen2\font plus
\BIBentryALTinterwordstretchfactor\fontdimen3\font minus
  \fontdimen4\font\relax}
\providecommand{\BIBforeignlanguage}[2]{{%
\expandafter\ifx\csname l@#1\endcsname\relax
\typeout{** WARNING: IEEEtran.bst: No hyphenation pattern has been}%
\typeout{** loaded for the language `#1'. Using the pattern for}%
\typeout{** the default language instead.}%
\else
\language=\csname l@#1\endcsname
\fi
#2}}
\providecommand{\BIBdecl}{\relax}
\BIBdecl

\bibitem{finlayson2021clinican}
``The clinician and dataset shift in artificial intelligence,'' \emph{The New
  England Journal of Medicine}, 2021.

\bibitem{fedavg}
B.~McMahan, E.~Moore, D.~Ramage, S.~Hampson, and B.~A. y~Arcas,
  ``Communication-efficient learning of deep networks from decentralized
  data,'' in \emph{Artificial intelligence and statistics}.\hskip 1em plus
  0.5em minus 0.4em\relax PMLR, 2017, pp. 1273--1282.

\bibitem{Hatamizadeh_2022_CVPR}
A.~Hatamizadeh, H.~Yin, H.~R. Roth, W.~Li, J.~Kautz, D.~Xu, and P.~Molchanov,
  ``Gradvit: Gradient inversion of vision transformers,'' in \emph{Proceedings
  of the IEEE/CVF Conference on Computer Vision and Pattern Recognition
  (CVPR)}, June 2022, pp. 10\,021--10\,030.

\bibitem{appfl}
M.~Ryu, Y.~Kim, K.~Kim, and R.~K. Madduri, ``{APPFL}: Open-source software
  framework for privacy-preserving federated learning,'' in \emph{2022 IEEE
  International Parallel and Distributed Processing Symposium Workshops
  (IPDPSW)}.\hskip 1em plus 0.5em minus 0.4em\relax IEEE, 2022, pp. 1074--1083.

\bibitem{beutel2020flower}
D.~J. Beutel, T.~Topal, A.~Mathur, X.~Qiu, J.~Fernandez-Marques, Y.~Gao,
  L.~Sani, K.~H. Li, T.~Parcollet, P.~P.~B. de~Gusm{\~a}o \emph{et~al.},
  ``Flower: A friendly federated learning research framework,'' \emph{arXiv
  preprint arXiv:2007.14390}, 2020.

\bibitem{roth2022nvidia}
H.~R. Roth, Y.~Cheng, Y.~Wen, I.~Yang, Z.~Xu, Y.-T. Hsieh, K.~Kersten,
  A.~Harouni, C.~Zhao, K.~Lu \emph{et~al.}, ``Nvidia flare: Federated learning
  from simulation to real-world,'' \emph{arXiv preprint arXiv:2210.13291},
  2022.

\bibitem{2022NatSD...9..657R}
N.~{Ravi}, P.~{Chaturvedi}, E.~A. {Huerta}, Z.~{Liu}, R.~{Chard},
  A.~{Scourtas}, K.~J. {Schmidt}, K.~{Chard}, B.~{Blaiszik}, and I.~{Foster},
  ``{FAIR principles for AI models with a practical application for accelerated
  high energy diffraction microscopy},'' \emph{Scientific Data}, vol.~9, no.~1,
  p. 657, Nov. 2022.

\bibitem{Huerta_2023_FAIRAI}
E.~A. Huerta, B.~Blaiszik, L.~C. Brinson, K.~E. Bouchard, D.~Diaz, C.~Doglioni,
  J.~M. Duarte, M.~Emani, I.~Foster, G.~Fox, P.~Harris, L.~Heinrich, S.~Jha,
  D.~S. Katz, V.~Kindratenko, C.~R. Kirkpatrick, K.~Lassila-Perini, R.~K.
  Madduri, M.~S. Neubauer, F.~E. Psomopoulos, A.~Roy, O.~Rübel, Z.~Zhao, and
  R.~Zhu, ``{FAIR} for {AI}: An interdisciplinary and international community
  building perspective,'' \emph{Scientific Data}, vol.~10, no.~1, jul 2023.

\bibitem{lamport2019byzantine}
L.~Lamport, R.~Shostak, and M.~Pease, ``The byzantine generals problem,'' in
  \emph{Concurrency: the works of leslie lamport}, 2019, pp. 203--226.

\bibitem{globus}
I.~Foster, ``Globus online: Accelerating and democratizing science through
  cloud-based services,'' \emph{IEEE Internet Computing}, vol.~15, no.~3, pp.
  70--73, 2011.

\bibitem{saas}
B.~Allen, J.~Bresnahan, L.~Childers, I.~Foster, G.~Kandaswamy, R.~Kettimuthu,
  J.~Kordas, M.~Link, S.~Martin, K.~Pickett \emph{et~al.}, ``Software as a
  service for data scientists,'' \emph{Communications of the ACM}, vol.~55,
  no.~2, pp. 81--88, 2012.

\bibitem{globusauth}
S.~Tuecke, R.~Ananthakrishnan, K.~Chard, M.~Lidman, B.~McCollam, S.~Rosen, and
  I.~Foster, ``Globus auth: A research identity and access management
  platform,'' in \emph{2016 IEEE 12th International Conference on e-Science
  (e-Science)}.\hskip 1em plus 0.5em minus 0.4em\relax IEEE, 2016, pp.
  203--212.

\bibitem{funcx}
R.~Chard, Y.~Babuji, Z.~Li, T.~Skluzacek, A.~Woodard, B.~Blaiszik, I.~Foster,
  and K.~Chard, ``{FuncX}: A federated function serving fabric for science,''
  in \emph{Proceedings of the 29th International symposium on high-performance
  parallel and distributed computing}, 2020, pp. 65--76.

\bibitem{fedavgm}
T.-M.~H. Hsu, H.~Qi, and M.~Brown, ``Measuring the effects of non-identical
  data distribution for federated visual classification,'' \emph{arXiv preprint
  arXiv:1909.06335}, 2019.

\bibitem{fedadaptive}
S.~Reddi, Z.~Charles, M.~Zaheer, Z.~Garrett, K.~Rush, J.~Kone{\v{c}}n{\`y},
  S.~Kumar, and H.~B. McMahan, ``Adaptive federated optimization,'' \emph{arXiv
  preprint arXiv:2003.00295}, 2020.

\bibitem{fedasync}
C.~Xie, S.~Koyejo, and I.~Gupta, ``Asynchronous federated optimization,''
  \emph{arXiv preprint arXiv:1903.03934}, 2019.

\bibitem{fedbuf}
J.~Nguyen, K.~Malik, H.~Zhan, A.~Yousefpour, M.~Rabbat, M.~Malek, and D.~Huba,
  ``Federated learning with buffered asynchronous aggregation,'' in
  \emph{International Conference on Artificial Intelligence and
  Statistics}.\hskip 1em plus 0.5em minus 0.4em\relax PMLR, 2022, pp.
  3581--3607.

\bibitem{mnist}
L.~Deng, ``The mnist database of handwritten digit images for machine learning
  research [best of the web],'' \emph{IEEE signal processing magazine},
  vol.~29, no.~6, pp. 141--142, 2012.

\end{thebibliography}
\end{document}